# Gel-OPTOFORT Sensor: Multi-axis Force/Torque Measurement and Geometry Observation Using GelSight and Optoelectronic Sensor Technology

*Yohan Noh (Brunel University London), Harshal Upare (Brunel University London),

Dalia Osman (Brunel University London), Wanlin Li (Beijing Institute for General Artificial Intelligence)

*Corresponding author (yohan.noh@brunel.ac.uk)

Although conventional GelSight-based tactile and force/torque sensors excel in detecting objects' geometry and texture information while simultaneously sensing multi-axis forces, their performance is limited by the camera's lower frame rates and the inherent properties of the elastomer. These limitations restrict their ability to measure higher force ranges at high sampling frequencies. Besides, due to the coupling of the Gelsight sensor unit and multi-axis force/torque unit structurally, the force/torque measurement ranges of the Gelsight-based force/torque sensors are not adjustable. To address these weaknesses, this paper proposes the GEL-OPTOFORT sensor that combines a GelSight sensor and an optoelectronic sensor-based force/torque sensor.

## 1. Introduction

In modern times, it is essential to embed various sensors into robotic applications to enhance the performance of AI (Artificial Intelligence). In particular, tactile and force/torque sensors facilitate robots' perception of touch with high intelligence when they interact with environments and objects.

In general, tactile and force/torque sensors have been fabricated using strain gauges, polyvinylidene fluoride (PVDF) films, piezoresistive sensors, and fiber Bragg grating (FBG), which are directly attached to an elastic body or its surface to measure deformation and estimate external force and torque components [1-3]. Alternatively, tactile and force/torque sensors can be designed using proximity sensing units such as capacitive sensors, magnets/hall sensors, optoelectronic sensors, etc. This approach indirectly measures the deformation of an elastic body by estimating distances to determine external force and torque components [4-6].

The issue is that while these tactile and force/torque sensors specialise in estimating the net force and torque, which are the sum of all force and torque components, they are incapable of measuring local force/torque components. Additionally, these sensors cannot perceive objects' geometry and texture information.

To overcome this issue, Gelsight-based tactile and force/torque sensors have been proposed to detect objects' geometry and texture information while simultaneously sensing multi-axis forces [7-10]. These Gelsight-based tactile and force/torque sensors consist of three units: 1) an elastomer unit with three groups of RGB LEDs; 2) a camera; 3) an elastic structure. The three groups of LEDs (red, green, blue) are positioned with an angle along the x-y planes on the outside of the elastomer in a radial direction with equal angles, and the top surface of an elastomer is coated with aluminum powder [7-9]. The elastomer unit with the three groups of RGB LEDs is mounted with a transparent acrylic plate (on which markers are printed) on the elastic structure [9-10]. This elastic structure secures a space in the centre, allowing a camera to be integrated into this space. When objects are pressed by the Gelsight sensor, the camera detects their geometry and texture information created by its captured image reflected by the three RGB LEDs. Furthermore, the camera detects the movement of markers' or ARTag's positions to estimate multi-axis force/torque components [9-10].

Although the Gelsight sensor integrating a multi-axis force/torque sensor has such remarkable multi-modal sensing capabilities, the following issues need to be resolved: 1) most miniature cameras capture images at 30 FPS, meaning the maximum sampling rate of tactile and force/torque sensing is limited to 33ms (the actual sampling rate is usually much lower than the maximum) [9-10]; 2) large deformation of the elastic structure should be required to sense large-range forces indirectly measured by the high-resolution marker or ARTag movement from the camera, resulting in significant errors due to the hysteresis of the structure for tactile and force/torque sensing [9-10]; 3) Although the elastic structure of the sensor [9-10] is replaceable with one of a different stiffness for adjusting different force/torque measurement ranges, it is still structurally coupled with the Gelsight unit, meaning that a force/torque calibration is needed after switching to the new elastic structure.

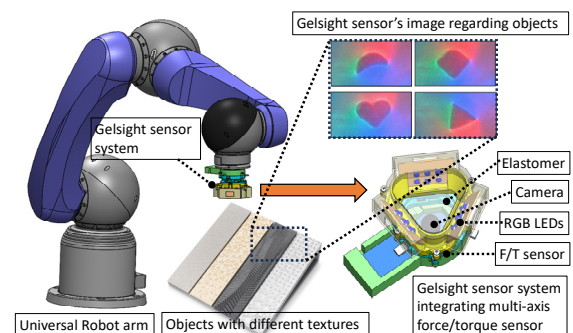

Fig. 1 Concept image of a Gel-OPTOFORT sensor

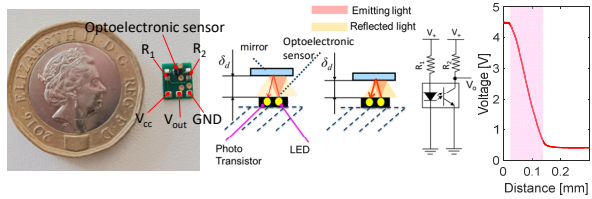

Fig. 2 Principle of sensing distances using an optoelectronic sensor

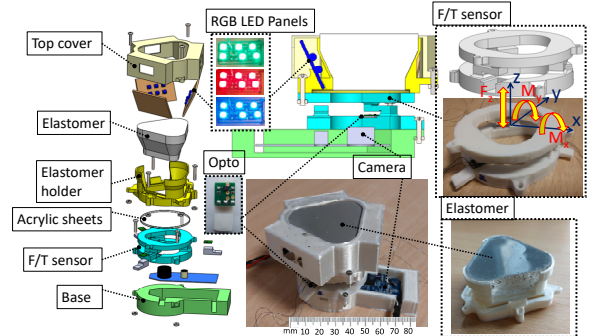

Fig. 3 Configuration of a Gel-OPTOFORT sensor

This causes a time-consuming calibration process.

Therefore, we propose a Gel-OPTOFORT (Gelsight sensor integrating OPTOelectronic sensor-based FORce/Torque) sensor, as shown in Fig. 1. Our proposed sensor has texture and force/torque detection modes that can operate either independently or collaboratively. This means the sensor can perceive an abundance of texture information in the environment at a low sampling frequency (for instance, classification tasks), provide precise force feedback for force servoing tasks at a high sampling frequency (for instance, force-feedback control), or deliver both high-resolution texture information and precise force information in scenarios that require both (for instance, grasping tasks).

Optoelectronic sensors are one of the proximity sensing units, and specialise in measuring high-precision distances with high-resolution and a high sampling rate without amplifiers (Figure 2) [11-13]. Additionally, the characteristic curve between the output of the voltage of optoelectronic sensors and distance shows very good linearity, and these sensors are immune to noise generated by electric and magnetic fields. Thus, without low-pass filters, their output voltages are relatively less noisy compared to the other proximity sensing units and direct attachment sensing units above mentioned (Figure 2) [11-13]. Low-cost miniature optoelectronic sensors are available on the market (approximately £0.5 per unit). By mounting optoelectronic sensors under reflective surfaces, distances can be easily measured. Once a PCB (printed circuit board) with mounted optoelectronic sensors is fabricated, the overall size of multi-axis force/torque sensors can be miniaturised, and the manufacturing process becomes simple, making this approach suitable for mass production [6,14-15]. As a result, overall manufacturing costs can be further reduced.

Therefore, in this paper, by leveraging the advantages of optoelectronic sensors, we present a low-cost modular Gel-OPTOFORT sensor which resolves the current issues of the Gelsight sensor integrating multi-axis force/torque sensing mentioned above.

## 2. Sensor design and fabrication

### 2.1 Design Concept

The proposed Gel-OPTOFORT sensor consists of two units such as a Gelsight sensor unit and a multi-axis force/torque sensor unit as shown in Fig. 3. For the Gelsight sensor unit, all the components such as RGB LEDs (Osram, Germany), a camera (C270 HD Webcam, 720p, Logitech, Switzerland), and an elastomer (Smooth-On, Inc., USA) were replicated as indicated in following published papers [7, 9]. As mentioned in the earlier, due to the coupling of the Gelsight sensor unit and multi-axis force/torque unit, calibration was conducted with the force/torque sensor unit integrated. If either one of the two units is fractured and replaced, calibration should be redone with the two units assembled again. Therefore, the design of the Gel-OPTOFORT sensor should be modularised to completely decouple the three units: the elastomer unit with RGB LEDs, the multi-axis force/torque sensor unit, and the camera unit.

### 2.2 Modular multi-axis force/torque sensor

The multi-axis force/torque sensor comprises three cantilever beams, and three optoelectronic sensors (NJL5901R-2, $1.0 \times 1.4 \times 0.6$ mm$^3$, New Japan Radio Co., Ltd., Tokyo, Japan), which are fixed under the beams, with reflectors attached to the underside of the structure, opposing the optoelectronic sensors. The elastic structure is fabricated by a 3D printer using PLA (Polylactic acid). This elastic structure allows for the measurement of one force $F_z$ and two moments $M_x$ and $M_y$ (Fig. 4). Assuming applied forces on the surface of an elastomer, $F_x$ and $F_y$ can be estimated based on $M_x$ and $M_y$. The three optoelectronic sensors can measure three deflections, thereby estimating three-axis force/torque components [16]. Optoelectronic sensors can measure tiny distances ranging from 0 to 0.1mm as depicted in Fig. 2 (purple area) [11-12]. Hence, the initial gap between the reflector (mirror) and the optoelectronic sensor should be 0.05 mm to accurately measure posi-tive/negative deflections relative to this initial gap.

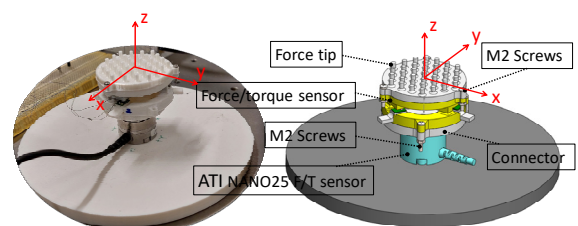

Fig. 4 Calibration platform for the proposed multi-axis F/T sensor using ATI NANO 25

The measurement range of force/torque sensing can be tailored by adjusting the cross-sectional area of length of the cantilever beam [16]. Finally, the designed elastic structure undergoes Finite Element Analysis (FEA) to confirm that the observed deflections remain within the specified distance ranges highlighted in the purple area when subjected to maximum force and torque values.

Since The geometry of the multi-axis force/torque sensor is a ring-like structure, this facilitates integration of a camera. The elastomer unit (yellow colour), multi-axis force/torque sensor unit (sky-blue colour), and camera unit (green colour) are simply assembled using screws as depicted in Fig. 3. Besides, the proposed multi-axis force/torque sensor can be independently used in various robot applications. The measurement range of force and torque sensing of the Gel-OPTOFORT sensor can be adjusted by integrating prefabricated force/torque sensors with different measurement ranges.

## 3. Multi-axis force/torque sensor calibration

$$\begin{bmatrix} F_z \\ M_x \\ M_y \end{bmatrix} = \begin{bmatrix} k_1 & k_2 & k_3 \\ k_4 & k_5 & k_6 \\ k_7 & k_8 & k_9 \end{bmatrix} \cdot \begin{bmatrix} v_1 \\ v_2 \\ v_3 \end{bmatrix} \quad (1)$$

$$\text{,where } k = \begin{bmatrix} -0.0201 & -0.0109 & -0.0267 \\ 3.2639e-04 & -1.0602e-04 & -1.4194e-04 \\ 0.1709e-04 & -2.4107e-04 & 6.6255e-04 \end{bmatrix}$$

Calibration is a process that converts the output voltages of electronic sensors into physical quantities. Here, the output voltages of the optoelectronic sensors need to be converted into force and torque values. Following the fastest calibration process established in previous papers [9, 12], a commercialised force/torque sensor, NANO 25 (ATI Industrial Automation, USA) was utilised, and Arduino Mega (10-bit ADC) was used to record the output voltages of the optoelectronic sensors. The proposed multi-axis force/torque sensor with a calibration tip was mounted on the NANO 25 (Figure 4). Various force/torque conditions were applied to the force tip, and during this process, both the output voltage set ($v_1$, $v_2$, $v_3$) and the external force/torque set ($F_z$, $M_x$, $M_y$) were recorded simultaneously. The recorded data set was utilised to calculate a calibration matrix $k$ through *Linear Regression Method* as shown in Eq. (1) [9, 12].

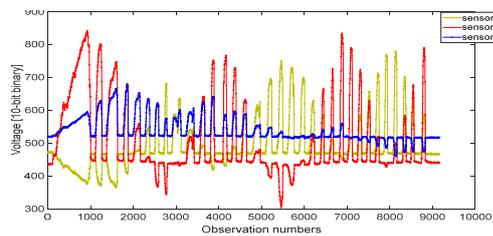

Fig. 5 The output voltages (10-bit ADC data) of three optoelectronic sensors

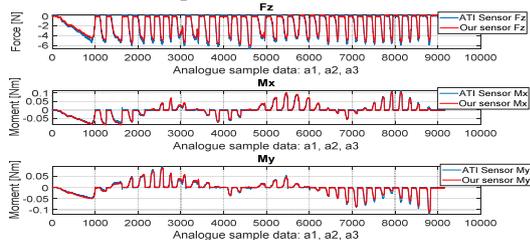

Fig. 6 The comparison between ground truth values and estimated force/torque ones

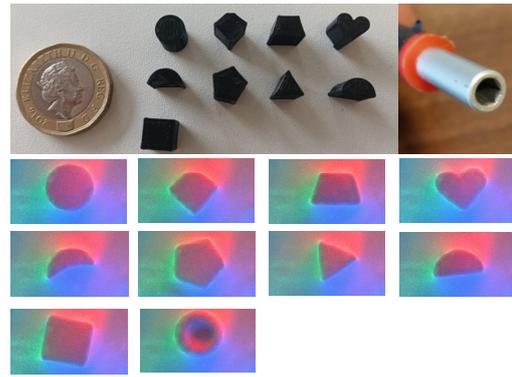

Fig. 7 Various testing objects and their Gelsight images

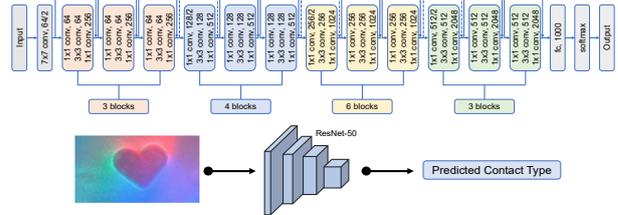

Fig. 8 ResNet architecture used in Gel-Optofort sensor

## 4. Experiments and results

### 4.1 Multi-axis Force/Torque sensor performance

The performance of the proposed multi-axis force/torque sensor was verified using the same calibration device as shown in Fig. 4. In the same way as the calibration process, a variety of external force/torque combinations were applied to the force tip. The estimated physical quantities were calculated by multiplying the obtained calibration matrix with the collected data. The estimated data were then compared with the ground truth data measured by the NANO 25 as shown in Figs. 5 and 6. The measurement ranges of the proposed force/torque sensor are $F_z = \pm 6$ N, $M_x = \pm 0.1$ N·m, and $M_y = \pm 0.1$ N·m, respectively. The root mean squared error (RMSE) of its measurements are 0.4391 N for $F_z$, 0.0051 N·m for $M_x$, and 0.0048 N·m for $M_y$, while the R-square values of its measurements are 0.9558 for $F_z$, 0.9730 for $M_x$, and 0.9648 for $M_y$.

### 4.2 Texture detection performance using Machine Learning algorithm

Images are captured using a Logitech C270 HD Webcam with a Python script and are transmitted to a host computer via USB 2.0. Images have a size of 1280 × 720 pixels and are effectively transferred at 30 fps. A ResNet-50 architecture is used as our machine-learning model for contact shape classification. The data for sin-gle contact includes a total of 30587 samples at 10 randomly selected shapes including circle, concentric circle, pentagon, semi-circle, square, diamond, heart, moon, trapezium, and triangle. The data set is split into training and validation subsets with a ratio of 80% (24469) and 20% (6118). We use the full four blocks of ResNet (Figure 8), comprising 50 layers in total, to estimate the contact shape directly. The machine-learning model is trained with a batch size of 64 for 5 epochs, using Adam optimiser with a learning rate of 0.001 for cross-entropy loss (CE) minimisation. The performance of the model achieves 99.9% accuracy in the classification task.

## 5. Conclusions and future works

In this paper, we have demonstrated the concept of the Gel-OPTOFORT sensor which decouples a Gelsight sensor and a multi-axis force/torque sensor (using optoelectronic sensors and a printed elastic structure) as shown in Fig. 3. The concept of the Gel-OPTOFORT sensor was verified through the preliminary experiments on the two sensing performance tests described in Chapter 4.

There are several issues to resolve to improve the accuracy of force/torque sensors. In general, the accuracy of force/torque sensors is affected by the material properties and geometry of an elastic structure. The elastic structure fabricated with PLA exhibits significant hysteresis, causing force/torque sensors to have lower repeatability and sensitivity in estimating force/torque values. Additionally, the current elastic structure geometry needs to be designed to mechanically decouple the estimated force/torque values ($F_x$, $F_y$ and $F_z$). Until these issues are sorted out, accuracy cannot be further improved. In the future, more detailed performance tests on a redesigned force/torque sensor fabricated with metal and a new geometry should be conducted. These tests should evaluate accuracy, hysteresis, repeatability, and linearity.

The Gelsight sensor unit was tested with a few object's shapes through the machine learning algorithm, but it should be tested with more complex object shapes and textures in future work. Additionally, possible scenarios using the Gel-OPTOFORT sensor will be investigated to determine how its texture and force/torque detection modes can be operated either independently or collaboratively, or both.

## ACKNOWLEDGEMENT

The research leading to these results has received funding from MSc project 2022-2023, Department of Mechanical and Aerospace Engineering, Brunel University London. Additionally, the travel budget to Japan for participating in the 42nd annual conference of the Robotics Society of Japan was supported by the Great Britain Sasakawa Foundation.